\documentclass[12pt]{amsart}
\usepackage{amssymb}
\usepackage{xspace}
\usepackage{url}
\usepackage{placeins}
\usepackage{needspace}
\usepackage{float}
\usepackage{todonotes}

\usepackage{pgf}
\usepackage{tikz}
\usetikzlibrary{arrows,automata}
\usetikzlibrary{positioning}

\tikzset{
    state/.style={
           rectangle,
           rounded corners,
           draw=black, thick,
           minimum height=2em,
           inner sep=2pt,
           text centered,
           },
}

\newcommand{\dfn}[1]{{\bf #1}}
\newcommand{\inference}[2]{\genfrac{}{}{1pt}{}{#1}{#2}}
\newcommand{\mymathop}[1]{\mathop{\texttt{#1}}}
\newcommand{\V}[1]{\ensuremath{\texttt{\mbox{#1}}}}
\newcommand{\type}[1]{\ensuremath{\texttt{#1}}}

\newcommand{\isWellDef}{\mathop{\downarrow}}
\newcommand{\isIllDef}{\mathop{\uparrow}}

\newcommand{\unicorn}{\ensuremath{\bot}\xspace}
\newcommand{\TRUE}{\ensuremath{\mathbb T}\xspace}
\newcommand{\FALSE}{\ensuremath{\mathbb F}\xspace}
\newcommand{\tuple}[1]{\ensuremath{\left\langle #1 \right\rangle}}

\newcommand{\seq}[1]{\ensuremath{\left[ #1 \right]}}

\newcommand{\typed}[2]{\ensuremath{#1\mcolon#2}}
\newcommand{\Vtyped}[2]{\ensuremath{\V{#1}\mcolon#2}}
\newcommand{\VTtyped}[2]{\ensuremath{\V{#1}\mcolon\type{#2}}}

\newcommand{\mname}[1]{\mbox{\sf #1}}
\newcommand{\Oname}[1]{\mname{#1}}
\newcommand{\OA}[2]{\ensuremath{%
    \mathop{\Oname{#1}}\mathopen{}\left(#2\right)\mathclose{}%
}}

\newcommand{\rawfn}[2]{\ensuremath{\mathop{#1}\mathopen{}\left(#2\right)\mathclose{}}}
\newcommand{\Fname}[1]{\V{#1}}
\newcommand{\FA}[2]{\ensuremath{\rawfn{\mathop{\Fname{#1}}}{#2}}}
\newcommand{\VFA}[2]{\FA{#1}{\V{#2}}}

\newcommand{\Hd}[1]{\OA{hd}{#1}}
\newcommand{\Tl}[1]{\Oname{tl}({#1})}
\newcommand{\HTl}[1]{\Hd{\Tl{#1}}}
\newcommand{\TTl}[1]{\Tl{\Tl{#1}}}
\newcommand{\HTTl}[1]{\Hd{\Tl{\Tl{#1}}}}
\newcommand{\TTTl}[1]{\Tl{\TTl{#1}}}

\newcommand{\defined}{\underset{\text{def}}{\equiv}}
\newcommand{\naturals}{\ensuremath{\mathbb{N}}\xspace}
\newcommand{\cat}{\mathbin{\scalebox{2}{.}}}
\newcommand{\expon}{\mathbin{{*}{*}}}
\newcommand{\mult}{\mathbin{\ast}}
\newcommand{\order}[1]{\ensuremath{{\mathcal O}(#1)}\xspace}

\newcommand{\size}[1]{\ensuremath{\left | {#1} \right |}}
\newcommand{\Vsize}[1]{\ensuremath{\size{\V{#1}}}}

\newcommand{\de}{\mathrel{::=}}
\newcommand{\derives}{\mathrel{\Rightarrow}}
\newcommand{\destar}
    {\mathrel{\mbox{$\stackrel{\!{\ast}}{\Rightarrow\!}$}}}
\newcommand{\deplus}
    {\mathrel{\mbox{$\stackrel{\!{+}}{\Rightarrow\!}$}}}
\newcommand{\mydot}{\raisebox{.05em}{$\,\bullet\,$}}

\newcommand{\lastix}[1]{\ensuremath{\##1}}
\newcommand{\Vlastix}[1]{\ensuremath{\lastix{\V{#1}}}}
\newcommand{\VlastElement}[1]{\Velement{#1}{\Vlastix{#1}}}

\newcommand{\Velement}[2]{\ensuremath{\mathop{\V{#1}}
    \mathopen{}\left[#2\right]\mathclose{}}}
\newcommand{\VVelement}[2]{\ensuremath{\mathop{\V{#1}}
    \mathopen{}\left[ \V{#2}
    \right]\mathclose{} }}

\newcommand{\mcolon}{\mathrel:}
\newcommand{\mcoloncolon}{\mathrel{\vcenter{\hbox{$::$}}}}

\newcommand{\suchthat}{\mcoloncolon}
\newcommand{\quantify}[3]{
    \ensuremath{\mathrel{#1}#2 \; \suchthat \; #3}%
}

\newcommand{\setB}[2]{\lbrace #1 \; \suchthat \; #2 \rbrace}

\newcommand{\existQOp}{\mathop{\exists{}}}
\newcommand{\existQ}[2]{\quantify{\existQOp}{#1}{#2}}

\newcommand{\fnMap}{\rightharpoonup}
\newcommand{\tfnMap}{\rightarrow}

\newcommand{\tfnMMap}{\ensuremath{
    \mathrel{\raise.75pt\hbox{\ensuremath{
          \raise1.19pt\hbox{\scalebox{.60}{$\ni$}} \mkern-12mu \rightarrow}}}
    }}

\newcommand{\Vah}[1]{\ensuremath{\V{#1}_{\type{AH}}}}

\newcommand{\Vdr}[1]{\ensuremath{\V{#1}_{\type{DR}}}}

\newcommand{\setOf}[1]{{{#1}^{\mbox{\normalsize $\ast$}}}}
\newcommand{\Vrule}[1]{\ensuremath{\V{#1}_{\type{RULE}}}}

\newcommand{\Veim}[1]{\ensuremath{\V{#1}_{\type{EIM}}}}
\newcommand{\Vahem}[1]{\ensuremath{\V{#1}_{\type{AHEM}}}}
\newcommand{\es}[1]{\ensuremath{#1_{\type{ES}}}}
\newcommand{\Ves}[1]{\ensuremath{\V{#1}_{\type{ES}}}}
\newcommand{\Vstr}[1]{\ensuremath{\V{#1}_{\type{STR}}}}

\newcommand{\Vsym}[1]{\ensuremath{\V{#1}_{\type{SYM}}}}

\newcommand{\Eloc}[1]{\ensuremath{{#1}_{\type{LOC}}}}
\newcommand{\Vloc}[1]{\Eloc{\V{#1}}}

\newcommand{\Cw}{\V{w}}

\newcommand{\Nulling}[1]{\mymathop{Nulling}(#1)}
\newcommand{\Nullable}[1]{\mymathop{Nullable}(#1)}
\newcommand{\GOTO}[2]{\mymathop{GOTO}(#1, #2)}
\newcommand{\Next}[1]{\mymathop{Next}(#1)}

\newcommand{\Postdot}[1]{\mymathop{Postdot}(#1)}
\newcommand{\Pos}[1]{\mymathop{Pos}(#1)}
\newcommand{\Rule}[1]{\mymathop{Rule}(#1)}
\newcommand{\LHS}[1]{\mymathop{LHS}(#1)}
\newcommand{\RHS}[1]{\mymathop{RHS}(#1)}
\newcommand{\Rules}[1]{\ensuremath{\mymathop{Rules}(#1)}}
\newcommand{\Vocab}[1]{\ensuremath{\mymathop{Vocab}(#1)}}
\newcommand{\Terminals}[1]{\ensuremath{\mymathop{Terminals}(#1)}}
\newcommand{\NT}[1]{\ensuremath{\mymathop{NT}(#1)}}
\newcommand{\Accept}[1]{\ensuremath{\mymathop{Accept}(#1)}}
\newcommand{\DR}[1]{\mymathop{DR}(#1)}
\newcommand{\AH}[1]{\mymathop{AH}(#1)}

\newcommand{\Origin}[1]{\mymathop{Origin}(#1)}
\newcommand{\Current}[1]{\mymathop{Current}(#1)}
\newcommand{\myL}[1]{\mymathop{L}(#1)}

\newcommand\Etable[1]{\ensuremath{\mymathop{table}[#1]}}
\newcommand\bigEtable[1]{\ensuremath{\mymathop{table}\bigl[#1\bigr]}}
\newcommand\Vtable[1]{\Etable{\V{#1}}}

\newsavebox{\myBox}
\newlength{\myHt}
\newlength{\myDp}

\newcommand{\padBoxC}[3]{%
    \savebox{\myBox}{\ignorespaces#3}%
    \usebox{\myBox}%
    \myHt=\ht\myBox%
    \myDp=\dp\myBox%
    \vrule height\dimexpr\myHt+#1 depth\dimexpr\myDp+#2 width0pt%
}

\newcommand{\cellboxWidth}[1]{4.5in}

\newcommand{\cellboxB}[2]{%
    \parbox{#1}{\ignorespaces#2}%
}

\newcommand{\cellbox}[2][\cellboxWidth]{%
    \cellboxB{#1}{%
        \padBoxC{4pt}{0pt}{\vphantom{(}}%
        {\ignorespaces#2}%
        \padBoxC{0pt}{4pt}{\vphantom{(}}%
    }%
}

\newenvironment{MYsloppy}[1][3em]{\par\tolerance9999 \emergencystretch#1\relax}{\par}

\begin{document}

\date{\today}

\title{Marpa and nullable symbols}

\author{Jeffrey Kegler}
\thanks{%
Copyright \copyright\ 2023 Jeffrey Kegler.  Version 1.
}
\thanks{%
This document is licensed under
a Creative Commons Attribution-NoDerivs 3.0 United States License.
}

\begin{abstract}
Marpa~\cite{Marpa2023} was intended
to make the best results
in the academic literature
on Earley's algorithm
available as
a practical general parser.
Earley-based parsers have had issues handling
nullable symbols.
Initially, we dealt with nullable symbols by following the approach
in Aycock and Horspool's 2002 paper~\cite{AH2002}.
This paper reports our experience with ~\cite{AH2002},
and the approach to handling nullables that we settled on
in reaction to that experience.
\end{abstract}

\maketitle

\section{Overview}

The Marpa recognizer~\cite{Marpa2023} was intended
to make the best results
in the academic literature
on Earley's algorithm
available as
a practical general parser.
Accordingly, when Marpa was first released in 2011,
it included the improved handling of nullable symbols
in Aycock and Horspool's 2002 paper~\cite{AH2002}.

Section \ref{sec:preliminaries}
deals with notation and other conventions.
Sections \ref{sec:earley}
and \ref{sec:earley-ops}
describe the Earley algorithm.
Sections \ref{sec:aycock-horspool-ideal-solution},
\ref{sec:aycock-horspool-finite-automata},
and \ref{sec:nihilist-normal-form}
describe the Aycock-Horspool version of Earley's
algorithm~\cite{AH2002}.
Sections \ref{sec:pitfall-of-counting-earley-items}
and \ref{sec:ahfa-states-are-not-disjoint}
raise some theoretical issues about \cite{AH2002}.
Section \ref{sec:problems-with-the-ahfa}
discusses the practical issues we encountered in
implementing~\cite{AH2002}.
In 2014 we extensively modified Marpa's usage of
the ideas from \cite{AH2002}.
Section
\ref{sec:marpa-s-approach-to-nullable-symbols}
describes the new approach
we took in Marpa to deal with nullable symbols.
Section \ref{sec:conclusion} summarizes our conclusions.

\section{Preliminaries}
\label{sec:preliminaries}

Readers should be familiar with Marpa~\cite{Marpa2023}
and with standard grammar notation.
It will also be useful to be familiar with Earley parsing,
and with Aycock and Horspool's 2002 paper~\cite{AH2002}.

We use the type system
of Farmer~2012~\cite{Farmer2012},
without needing most of its apparatus.\footnote{%
Types in \cite{Farmer2012} are
collections of classes (``superclasses'').
But in this paper, every explicitly stated type will be a ZF set.
}
Let \V{exp} be an expression,
and let \type{T} be its type.
Type may be indicated in a ``wide'' notation:
\[ \VTtyped{exp}{T} \]
More often, this paper will use the ``narrow''
notation,
where a subscripts indicates type:
\[ \V{exp}_\type{T} \]

A noteworthy feature we adopt from Farmer~2012~\cite{Farmer2012} is his notion of ill-definedness.
For example, the value of partial functions may be ill-defined for
some arguments in their domain.
Farmer's handling of ill-definedness is the traditional one,
and was well-entrenched,
but he was first to describe and formalize it.\footnote{%
See Farmer 2004~\cite{Farmer2004}.
Note that Farmer refers to ill-defined values as ``undefined''.
We found this problematic.
For example, a partial function may not have a value for
every argument in its domain.
Saying that the value of the partial function for these arguments
is defined as ``undefined'' is confusing.
In this paper we say that all the values of partial functions
are defined, but that some may not have values in the codomain,
and are therefore ill-defined.
}

We write \TRUE for ``true''; \FALSE for ``false'';
and $\unicorn$ for ill-defined.
A value is \dfn{well-defined} iff it is not ill-defined.
We write $\V{x}\isWellDef$ to say that \V{x} is well-defined,
and we write $\V{x}\isIllDef$ to say that \V{x} is ill-defined.

Traditionally, and in this paper,
any formula with an ill-defined operand is false.
This means that an equality both of whose operands are ill-defined
is false, so that $\neg(\unicorn = \unicorn)$.
For cases where this is inconvenient,
we introduce a new relation, $\simeq$, such that
$$
  \V{a} \simeq \V{b} \defined \V{a} = \V{b} \lor (\V{a} \isIllDef \land \V{b} \isIllDef ).
$$

We often abbreviate ``if and only if'' to ``iff''.
We also often substitute the more prominent double colon ($\mcoloncolon$)
for the ``mid'' divider ($\mid$).

We define the natural numbers, \naturals, to include zero.
$\V{f} \circ \V{g}$, pronounced ``\V{f} after \V{g}'',
indicates the composition of the functions \V{f} and \V{g},
so that
$\rawfn{(\V{f}\circ\V{g})}{\V{x}} = \VFA{f}{\VFA{g}{\V{x}}}$.
The difference of the two sets, \V{S1} and \V{S2}, is written
\( \V{S1} \setminus \V{S2}. \)

We write tuples using angle brackets: $ \tuple{\V{a},\V{b},\V{c}} $.
The head of a tuple \V{S} can be written \Hd{S}.
The tail of a tuple \V{S} can be written \Tl{S}.
For example, where
\[
    \V{S} = \tuple{ 42, 1729, 42 },
\]
then \V{S} is a 3-tuple, or triple,
\begin{gather*}
\Hd{S}=42, \\
\Tl{S}=\tuple{1729,42}, \\
\HTl{S}=1729, \\
\TTl{S}=42, \text{ and} \\
\TTTl{S}\isIllDef.
\end{gather*}

We define the natural numbers, \naturals, to include zero.
$\V{D} \fnMap \V{C}$ is the set of partial functions from domain \V{D} to codomain \V{C}.
$\V{D} \tfnMap \V{C}$ is the set of total functions from domain \V{D} to codomain \V{C}.
It follows that $\naturals \tfnMap \V{C}$ is the set of
infinite sequences of terms from the set \V{C};
and that $42 \tfnMap \V{C}$ is the set of
sequences of length 42 of terms from the set \V{C}.
We say that
$$ \V{domSet} \tfnMMap \V{C} \defined
   \bigl\{ \V{fn} \bigm|
       \left( \exists \; \V{D} \in \V{domSet} \; \middle| \; \V{fn} \in \V{D} \tfnMap \V{C}
       \right) \bigr\}
$$
so that $\naturals \tfnMMap \V{C}$
is the set of finite sequences of terms from the set \V{C}.

We write \Vsize{\V{seq}} for the cardinality, or length,
of the sequence \V{seq}.
\VVelement{seq}{i} is the \V{i}'th term of the sequence \V{seq},
and is well-defined when
$0 \le \V{i} < \Vsize{seq}$.
We often specify a sequence by giving its terms inside
brackets.
For example,
$$\seq{\, \V{a}, \; 42, \; \seq{} \, }$$
is the sequence of length
3 whose terms are, in order, \V{a}, the number 42, and the empty sequence.

The last index of the sequence \V{seq} is \Vlastix{seq},
so that \VlastElement{seq} is the last term of \V{seq}.
\Vlastix{seq} is ill-defined for the empty sequence,
that is, if $\Vsize{seq} = 0$.
If \V{seq} is not the empty sequence, then
$\Vsize{seq} = \Vlastix{seq} + 1$.

Let \type{SYM} be the type for symbols, which
will will treat as opaque,
except that
\( \emptyset \notin \type{SYM} \).
The string type, \type{STR}, is the set of all finite sequences of
symbols:
\[ \type{STR} = \naturals \tfnMMap \type{SYM}. \]
The empty string is \( \emptyset \), but we let
\( \emptyset = \epsilon \)
and we usually write the empty string as \( \epsilon \).
We also use \(\epsilon\) as
the reserved non-symbol: \( \epsilon \notin \type{SYM} \).

A rule (type \type{RULE}) is a duple:
\[ \type{RULE} = \type{SYM} \times \type{STR}. \]
\[ \begin{gathered}
    \LHS{\VTtyped{r}{RULE}} = \Hd{\V{r}}. \\
    \RHS{\VTtyped{r}{RULE}} = \Tl{\V{r}}.
\end{gathered} \]
We usually write a rule \V{r} in
the form $\tuple{\Vsym{lhs} \de \Vstr{rhs}}$,
where $\V{lhs} = \LHS{\V{r}}$
and $\V{rhs} = \RHS{\V{r}}$.
\Vsym{lhs} is referred to as the left hand side (LHS)
of \Vrule{r}.
\Vstr{rhs} is referred to as the right hand side (RHS)
of \Vrule{r}.

A grammar is a 4-tuple, type \type{G},
\[
      \type{G} = \setOf{\type{SYM}} \times \setOf{\type{SYM}} \times
	\setOf{\type{RULE}} \times \type{SYM},
\]
where, if \VTtyped{g}{G} is a grammar, then
\[ \begin{gathered}
      \Vocab{\V{g}} \defined \Hd{\V{g}}, \\
      \Terminals{\V{g}} \defined \HTl{\V{g}}, \\
      \Terminals{\V{g}} \subset \Vocab{\V{g}}, \\
      \NT{\V{g}} \defined \Vocab{\V{g}} \setminus \Terminals{\V{g}}, \\
      \Rules{\V{g}} \defined \HTTl{\V{g}}, \\
      \V{r} \in \Rules{\V{g}} \implies \LHS{\V{r}} \in \NT{\V{g}}, \\
      \V{r} \in \Rules{\V{g}} \implies \RHS{\V{r}} \in (\naturals \tfnMMap \Vocab{\V{g}}), \\
      \Accept{\V{g}} \defined \TTTl{\V{g}},
	\text{ and }\\
      \Accept{\V{g}} \in \NT{\V{g}}. \\
\end{gathered} \]

The rules of a grammar imply the traditional rewriting system,
in which
\begin{itemize}
\item $\Vstr{x} \derives \Vstr{y}$
states that \V{x} derives \V{y} in exactly one step;
\item $\Vstr{x} \deplus \Vstr{y}$
states that \V{x} derives \V{y} in one or more steps;
and
\item $\Vstr{x} \destar \Vstr{y}$
states that \V{x} derives \V{y} in zero or more steps.
\end{itemize}
We call these rewrites \dfn{derivation steps}.
A sequence of zero or more derivation steps,
in which the left hand side of all but the first
is the right hand side of its predecessor,
is a \dfn{derivation}.
We say that the symbol \V{x} \dfn{induces}
the string of length 1 whose only term is that symbol,
that is, the string \seq{\Vsym{x}}.
Pedantically, the terms of derivations,
and the arguments of concatenations like
$\Vstr{s1} \cat \Vstr{s2}$,
must be strings.
But in concatenations and derivation steps we often
write the symbol to represent the string it induces so
that
$$ \Vstr{a} \cat \Vsym{b} \cat \Vstr{c}
= \Vstr{a} \cat \seq{\Vsym{b}} \cat \Vstr{c}.
$$

A \dfn{sentence} of \V{g} is a string of terminals derivable
from \Accept{\V{g}}.
The language of a grammar \V{g}, $\myL{\V{g}}$,
is its set of sentences:
\[ \begin{gathered}
  \myL{\V{g}} = \setB{ \VTtyped{sentence}{STR} }{ \\
  \V{sentence} \in \setOf{\Terminals{\V{g}}} \land  \Accept{\V{g}} \destar \V{sentence} }
\end{gathered} \]

We say that a string \V{x} is \dfn{nullable},
$\Nullable{\Vstr{x}}$, iff the empty string can be
derived from it:
$$\Nullable{\Vstr{x}} \defined \V{x} \destar \epsilon.$$
We say that a string \V{x} is \dfn{nulling},
$\Nulling{\Vstr{x}}$,
iff it always eventually derives the null
string:
\[
\Nulling{\Vstr{s}} \defined
  \forall \; \Vstr{y} \;\mid\; \V{x} \destar \V{y} \implies \V{y} \destar \epsilon.
\]
We say that symbols are nulling or nullable based on the string they
induce:
\begin{gather*}
\Nullable{\Vsym{x}} \defined \Nullable{\seq{\V{x}}}. \\
\Nulling{\Vsym{x}} \defined \Nulling{\seq{\V{x}}}.
\end{gather*}
A string or symbol is
\begin{itemize}
\item \dfn{non-nullable} iff it is not nullable;
\item \dfn{properly nullable} iff it is nullable,
but not nulling; and
\item \dfn{non-nulling} iff it is not nulling.
\end{itemize}
We often refer to nullable symbols as \dfn{nullables},
and to properly nullable symbols as \dfn{proper nullables},

We assume the grammars in this paper are ``augmented''.
We say that a grammar \V{g} is \dfn{augmented} iff
\begin{itemize}
\item there is an \dfn{accept rule},
\begin{equation}
\label{eq:accept-rule}
  \Vrule{accept} = \tuple{ \Vsym{accept} \de \Vsym{start} },
\end{equation}
where \( \Vsym{start} \in \Vocab{\V{g}} \)
and \( \Vsym{accept} = \Accept{\V{g}} \) is the \dfn{accept symbol};
\item the accept symbol does not appear as a RHS symbol in any rule,
\begin{equation*}
\begin{gathered}
\forall \; \V{x} \in \Rules{\V{g}} \mid
\; \nexists \; \Vstr{pre}, \; \Vstr{post} \; \mid \\
\V{pre} \cat \Vsym{accept} \cat \V{post} = \RHS{\Vrule{x}}; \text{ and} \\
\end{gathered}
\end{equation*}
\item the accept rule is unique,
\[ \Vsym{accept} = \LHS{\V{x}} \implies \Vrule{accept} = \V{x}. \]
\end{itemize}

Let the input to
the parse be $\VTtyped{w}{STR}$.
Locations in the input will be of type \type{LOC},
where \( \type{LOC} = \naturals \).
When we state our complexity results later,
they will often be in terms of $\V{n}$,
where $\V{n} = \Vsize{w}$.
\typed{\VVelement{w}{i}}{\type{SYM}} is the \Vloc{i}'th
character
of the input,
and is well-defined when
$0 \le \Vloc{i} < \Vsize{w}$.

\section{Earley's algorithm}
\label{sec:earley}

A dotted rule (type \type{DR}) is a duple,
\[ \type{DR} = \type{RULE} \times \naturals, \]
such that, if \V{dr} is of type \type{DR},
\[ \begin{gathered}
  \Rule{\V{dr}} \defined \Hd{\V{dr}}, \\
  \Pos{\V{dr}} \defined \Tl{\V{dr}}, \text{ and} \\
  \Pos{\V{dr}} \le \size{\Rule{\V{dr}}}.
\end{gathered} \]
We say that $\Rule{\V{dr}}$ is the \dfn{rule} of \V{dr}
and $\Pos{\V{dr}}$ is the \dfn{dot position} of \V{dr}.

The dot position of a dotted rule indicates the extent to which
the rule has been recognized,
and is represented with a large raised dot,
so that if
\begin{equation*}
\tuple{ \Vsym{A} \de \Vsym{X} \cat \Vsym{Y} \cat \Vsym{Z}}
\end{equation*}
is a rule,
\begin{equation*}
\tuple{ \Vsym{A} \de \V{X} \cat \V{Y} \mydot \V{Z}}
\end{equation*}
is the dotted rule with the dot at
$\V{pos} = 2$,
between \Vsym{Y} and \Vsym{Z}.

Every rule concept, when applied to a dotted rule,
is applied to the rule of the dotted rule.
The following are examples:
\[
\begin{gathered}
  \LHS{\Vdr{x}} \defined \LHS{\Rule{\V{x}}}. \\
  \RHS{\Vdr{x}} \defined \RHS{\Rule{\V{x}}}.
\end{gathered}
\]

We also make the following definitions:
\[
  \Postdot{\Vdr{x}} \defined
  \begin{cases}
  \Vsym{next}, \text{ if } \existQ{
      \Vstr{pre}, \; \Vstr{post}, \; \Vsym{A}}{ \\
      \qquad \V{x} = \tuple{ \V{A} \de \V{pre} \mydot \V{next} \cat \V{post}}}, \\
  \unicorn, \text{ otherwise.}
  \end{cases} \\
\]

\[
  \Next{\Vdr{x}} \defined
  \begin{cases}
    \VTtyped{\tuple{ \Vsym{A} \de \Vstr{pre} \cat \Vsym{next} \mydot \Vstr{post}}}{DR}, \\
      \qquad \text{if } \existQ{
      \Vstr{pre}, \; \Vstr{post}, \; \Vsym{A}}{ \\
      \qquad \qquad \V{x} = \tuple{ \V{A} \de \V{pre} \mydot \V{next} \cat \V{post}} }, \\
    \unicorn, \text{ otherwise.}
  \end{cases} \\
\]

The \dfn{initial dotted rule} is
\begin{equation}
\label{eq:initial-dr}
\Vdr{initial} = \tuple{\Vsym{accept} \de \mydot \Vsym{start} },
\end{equation}
where \Vsym{accept} and \Vsym{start} are as in
the accept rule,
\eqref{eq:accept-rule}
on page \pageref{eq:accept-rule}.
A \dfn{predicted dotted rule} is a dotted rule,
other than the initial dotted rule,
with a dot position of zero,
for example,
\begin{equation*}
\Vdr{predicted} = \tuple{\Vsym{A} \de \mydot \Vstr{alpha} }.
\end{equation*}
A \dfn{confirmed dotted rule}
is the initial dotted rule,
or a dotted rule
with a dot position greater than zero.
A \dfn{completed dotted rule} is a dotted rule with its dot
position after the end of its RHS,
for example,
\begin{equation*}
\Vdr{completed} = \tuple{\Vsym{A} \de \Vstr{alpha} \mydot }.
\end{equation*}
Predicted, confirmed and completed dotted rules
are also called, respectively,
\dfn{predictions}, \dfn{confirmations} and \dfn{completions}.

An Earley item (type \type{EIM}) is a triple,\footnote{%
  This definition of \type{EIM} departs from tradition.
  According to the tradition, the current location is not an element
  of the tuples which define \type{EIM}s.
  Instead \type{EIM}s
  are grouped into sets
  that share the same current location.
  These sets of co-located \type{EIM}s are called
  ``Earley sets''.
  Membership in an Earley set becomes a ``property''
  of each \type{EIM}.

  In set theory, by the Axiom of Extensionality,
  sets with the same membership (or extension)
  are equivalent.
  In set theory, non-extensional properties,
  such as membership in an Earley set,
  do not make sets with the same extension distinct.
  It can happen that \type{EIM}s in different Earley sets,
  by the Axiom of Extensionality, are equal as sets.
  Since in this paper, as in most modern mathematics, equality
  means equality as sets,
  and since \type{EIM}s at different current locations
  are conceptually distinct for almost all purposes,
  the omission of current location from the \type{EIM}
  definition is awkward.
  For this reason,
  in this paper, we honor the traditional definition of an \type{EIM}
  in the breach.
}
\[ \type{EIM} = \type{DR} \times \type{LOC} \times \type{LOC}, \]
such that, when \V{x} is of type \type{EIM},
\begin{gather*}
     \DR{\V{x}} \defined \Hd{\V{x}}, \\
     \Origin{\V{x}} \defined \HTl{\V{x}}, \text{ and} \\
     \Current{\V{x}} \defined \TTl{\V{x}}.
\end{gather*}
\begin{MYsloppy}
We say that $\DR{\Veim{x}}$ is the \dfn{dotted rule} of \V{x}.
We say that $\Current{\Veim{x}}$
is the \dfn{current location} of \Veim{x}.
The current location of an Earley item \Veim{x} is the location in the input
where the rule $\Rule{\DR{\Veim{x}}}$
was recognized as far as the dot in the dotted rule of \Veim{x}.
\end{MYsloppy}

We say that
$\Origin{\Veim{x}}$
is the \dfn{origin} of \Veim{x}.
The origin of an Earley item
is the location in the input where recognition of $\Rule{\DR{\Veim{x}}}$
started.
For convenience, the type \type{ORIG} will be a synonym
for \type{LOC}, indicating that the variable designates
the origin entry of an Earley item.

We find it convenient to apply dotted rule concepts to
\type{EIM}'s,
so that the concept applied to the \type{EIM} is
the concept applied to the dotted rule of the \type{EIM}.
The following are examples:
\begin{gather*}
\LHS{\Veim{x}} \defined \LHS{\DR{\V{dr}}}. \\
\RHS{\Veim{x}} \defined \RHS{\DR{\V{dr}}}. \\
\Pos{\Veim{x}} \defined \Pos{\DR{\V{dr}}}. \\
\Postdot{\Veim{x}} \defined \Postdot{\DR{\V{dr}}}. \\
\Next{\Veim{x}} \defined \Next{\DR{\V{dr}}}. \\
\Rule{\Veim{x}} \defined \Rule{\DR{\V{dr}}}.
\end{gather*}

An Earley parser builds a table of Earley sets,
\begin{equation*}
\Vtable{i},
\quad \text{where} \quad
0 \le \Vloc{i} \le \size{\Cw}.
\end{equation*}
Earley sets (type \type{ES}) are set of \type{EIM}s,
\[ \type{ES} = \setOf{\type{EIM}}. \]
Earley sets are often named by their location.
That is,
the bijection between \Ves{i} and \Vloc{i}
allows locations to be treated as the ``names'' of Earley sets.
We often write \Ves{i} to mean the Earley set at \Vloc{i},
and \Vloc{x} to mean the location of Earley set \Ves{x}.
The type designator \type{ES} is often omitted to avoid clutter,
especially in cases where the Earley set is not
named by location.
Occasionally the naming location is a expression,
so that
$$ \es{(\Origin{\Veim{x}})} $$
is the Earley set at the origin of the \type{EIM} \Veim{x}.
If \es{\V{working}} is an Earley set,
$\size{\es{\V{working}}}$ is the number of Earley items.

Recalling the accept rule,
\eqref{eq:accept-rule}
on page \pageref{eq:accept-rule},
we say that
the input \Cw{} is accepted if and only if
\begin{equation*}
\tuple{\tuple{\Vsym{accept} \de \Vsym{start} \mydot}, 0} \in \bigEtable{\Vsize{\Cw}}.
\end{equation*}

\section{Operations of the Earley algorithm}
\label{sec:earley-ops}

\textbf{Initialization}:
\begin{equation}
\label{eq:initial}
\inference{
   \TRUE
}{
    \begin{array}{c}
        \VTtyped{ \tuple{ \Vdr{initial}, 0, 0 }}{EIM} \\
    \end{array}
}
\end{equation}
Here \Vdr{initial} is
from \eqref{eq:initial-dr} on \pageref{eq:initial-dr}.
Earley {\bf initialization} only takes
place in Earley set 0,
and always adds exactly one \type{EIM}.

\vspace{1ex}
\textbf{Scanning}:
\begin{equation}
\label{eq:scan}
\inference{
    \begin{array}{c}
	\Postdot{\Veim{mainstem}} = \Cw\bigl[ \Vloc{current} \bigr]
    \end{array}
}{
    \begin{array}{c}
	\VTtyped{ \tuple{ \Next{\Veim{mainstem}}, \Origin{\V{mainstem}},
	    \V{current}+1
	}}{EIM}
    \end{array}
}
\end{equation}

\vspace{1ex}
\textbf{Reduction}:
\begin{equation}
\label{eq:reduction}
\inference{
    \begin{array}{c}
      \Current{\Veim{mainstem}} = \Origin{\Veim{tributary}} \\
      \Postdot{\Veim{mainstem}} = \LHS{\Veim{tributary}}
    \end{array}
}{
    \begin{array}{c}
	\left< \begin{gathered}
	  \Next{\Veim{mainstem}}, \Origin{\V{mainstem}}, \\
	       \Current{\V{tributary}}
	\end{gathered} \right> : \type{EIM}
    \end{array}
}
\end{equation}

\vspace{1ex}
\textbf{Prediction}:
\begin{equation}
\label{eq:prediction}
\inference{
    \begin{array}{c}
	\tuple{ \Vsym{lhs} \de \Vstr{rhs} } \in \Rules{\V{g}} \\
	\Postdot{\V{mainstem}} = \Vsym{lhs}
    \end{array}
}{
  \begin{array}{c}
      \left< \begin{gathered}
      \tuple{ \Vsym{lhs} \de \mydot \Vstr{rhs} }, \Current{\V{mainstem}}, \\
	  \Current{\V{mainstem}}
      \end{gathered} \right> : \type{EIM}
  \end{array}
}
\end{equation}

In traditional implementations,
the operations are applied
to create Earley sets,
in order from 0 to \Vsize{w}.
Duplicate \type{EIM}s are not added.
In typical implementations,
scanning is run
ahead of the other operations
in the sense that, while the other operations are adding
items to the Earley set at \Vloc{i},
scanned items are added to \es{(\V{i}+1)}.

Traditionally, each Earley set is implementated as a list.
The result of a prediction operation,
\eqref{eq:prediction} on page \pageref{eq:prediction},
may be the \V{mainstem} of a reduction operation,
\eqref{eq:reduction} on page \pageref{eq:reduction}.%
\label{text:prediction-completion-cycle}
That reduction operation may, in turn,
produce new predictions.
Typically, an implementation dealt with this
by making repeated passes
through the Earley set,
terminating when no more \type{EIM}s could be added.

\section{The Aycock-Horspool ``ideal'' solution}
\label{sec:aycock-horspool-ideal-solution}

The need to make multiple passes over each Earley set
was long seen as a problem.
Aycock and Horspool experimented with a suggestion
from Jay Earley that required
a dynamically-updated data structure,
but found this solution unsatisfactory~\cite[p. 621]{AH2002}.
Instead, they found a way to revise the prediction step
itself that eliminated the problem.

To present Aycock and Horspool's revised prediction step,
we first rewrite the original prediction operation
\eqref{eq:prediction} on page \pageref{eq:prediction},
in the form of a function,
\[ \Vtyped{Opred}{\type{EIM} \tfnMap \setOf{\type{EIM}}}, \]
such that
\begin{equation}
\label{eq:original-earley-prediction}
\begin{gathered}
\VFA{Opred}{\Veim{x}} =
\left\lbrace
    \Veim{pred} \; \middle| \;
    \begin{gathered}
     \LHS{\V{pred}} = \Postdot{\V{eim}} \\
     \land \, \Rule{\V{pred}} \in \Rules{\V{g}} \\
     \land \, \Pos{\V{pred}} = 0 \\
     \land \, \Current{\V{pred}} = \Current{\Veim{x}} \\
     \land \, \Origin{\V{pred}} = \Current{\Veim{x}}
     \end{gathered}
\right\rbrace . \end{gathered}
\end{equation}

Let
\( \Vtyped{AHpred}{\type{EIM} \tfnMap \setOf{\type{EIM}}} \)
be the transitive closure of \Fname{Opred}.
More precisely,
\begin{equation}
\label{eq:ah-earley-prediction}
\VFA{AHpred}{\VTtyped{x}{EIM}} = \Fname{predAll} \circ \Fname{Opred},
\end{equation}
where
\begin{equation*}
\begin{gathered}
\VFA{predAll}{\Vtyped{eimSet}{\setOf{\type{EIM}}}} =  \\
  \setB{\VTtyped{eim}{EIM}}{ \existQ{\Vtyped{i}{\naturals}}{
     \V{eim} \in \VFA{predN}{\V{i}, \V{eimSet}} } }, \\[1ex]
\VFA{predN}{0, \Vtyped{eimSet}{\setOf{\type{EIM}}}} = \V{eimSet}, \text{ and} \\[1ex]
\VFA{predN}{(\Vtyped{n}{\naturals})+1, \Vtyped{eimSet}{\setOf{\type{EIM}}}} = \\
  \setB{ \VTtyped{eim}{EIM} }{ \existQ{\VTtyped{eim2}{EIM}}{ \\
    \V{eim} \in \VFA{Opred}{\V{eim2}} \\
    \land \, \V{eim2} \in \VFA{predN}{\V{n},\V{eimSet}}}}.
\end{gathered}
\end{equation*}

Aycock and Horspool proved \cite[pp. 621-622]{AH2002} that,
by replacing the Earley prediction operation
(equation \ref{eq:original-earley-prediction}
on page \pageref{eq:original-earley-prediction}),
with its transitive closure
(equation \ref{eq:ah-earley-prediction}
on page \pageref{eq:ah-earley-prediction}),
they had created an algorithm that successful dealt with
nullable symbols.
Their algorithm required only one pass,
``retain[ed] the elegance of Earley's algorithm'',
and did not require a new, dynamically-updated data structure.

\section{The Aycock-Horspool finite automata}
\label{sec:aycock-horspool-finite-automata}

Aycock and Horspool called the algorithm of
the previous section ``ideal'',
but the quote marks are theirs
\cite[p. 621]{AH2002}.
This seems to reflect a perception on their part that,
while they had an idea
that allowed an Earley-based parse engine
to handle nullable symbols gracefully,
a practical implementation would demand some refinements.

The ``ideal'' solution of~\cite{AH2002},
if naively implemented,
required every prediction operation to compute a transitive closure.
But, during the parse, this transitive closure was a constant ---
it depended only on the grammar and the postdot symbol of the argument EIM.
Clearly, precomputation could eliminate most or all of the runtime
cost of their new prediction operation.

For their precomputation, Aycock and Horspool~\cite{AH2002}
invented a new semi-deterministic finite automata,
which \cite{AH2002} calls a ``split LR(0) $\epsilon$-DFA''.
In this paper,
we calling their ``split LR(0) $\epsilon$-DFA'',
an Aycock-Horspool Finite Automata (AHFA).

The AHFA is based
on a few observations.
\begin{itemize}
\item
In practice, Earley items sharing the same origin,
but having different dotted rules,
often appear together in the same Earley set.
\item
There is in the literature a method
for associating groups of dotted rules that often appear together
when parsing.
This method is the LR(0) DFA used in the much-studied
LALR and LR parsers.
\item
The LR(0) items that are the components of LR(0)
states are, exactly, dotted rules.
\item
By taking into account symbols that derive the
null string, the LR(0) DFA could be turned into an
LR(0) $\epsilon$-DFA,
which would be even more effective
at grouping dotted rules that often occur together
into a single DFA state.
\end{itemize}

Aycock and Horspool realized that,
by changing Earley items to track AHFA states
instead of individual dotted rules,
the size of Earley sets could be reduced,
and conjectured that this would
make Earley's algorithm faster in practice.

An AHFA state (type \type{AH}) is, in effect, a shorthand
for groups of dotted rules that occur together frequently,
\[ \type{AH} = \setOf{\type{DR}}. \]
We recall that
the traditional Earley items (\type{EIM}'s)
are triples, such that
\[ \VTtyped{x}{EIM} = \tuple{ \DR{\V{x}}, \Origin{\V{x}}, \Current{\V{x}} }, \]
where \Vdr{x} is a dotted rule.

An Aycock-Horspool Earley item (type \type{AHEM}) is a triple,
\[ \type{AHEM} = \type{AH} \times \type{ORIG} \times \type{LOC}, \]
such that, when \V{x} is of type \type{AHEM},
\begin{gather*}
     \AH{\V{x}} \defined \Hd{\V{x}}, \\
     \Origin{\V{x}} \defined \HTl{\V{x}}, \text{ and} \\
     \Current{\V{x}} \defined \TTl{\V{x}}.
\end{gather*}
We say that $\AH{\Vahem{x}}$ is the AHFA state of \V{x}.
Similarly to \type{EIM}s,
we say that $\Current{\Vahem{x}}$
is the \dfn{current location} of \Vahem{x},
and that
$\Origin{\Vahem{x}}$
is the \dfn{origin} of \Vahem{x}.

AHFA's are not fully deterministic --- they have
\dfn{null transitions}.
\cite{AH2002} defines
a partial transition function for
pairs of AHFA state and symbol,
\begin{equation*}
\Vtyped{GOTO}{ \type{AH} \times (\epsilon \cup \Vocab{\V{g}}) \fnMap \type{AH} }.
\end{equation*}
\begin{MYsloppy}
\noindent Null transitions are written as transitions on $\epsilon$,
the reserved non-symbol,
for example
\( \GOTO{\Vah{from}}{\epsilon}. \)
\end{MYsloppy}

If \Vah{predicted} is the result of a null transition,
it is called a \dfn{predicted} AHFA state.
If an AHFA state is not a \dfn{predicted} AHFA state,
it is called a \dfn{confirmed} AHFA state.
The initial AHFA state is a confirmed AHFA state.\footnote{%
In~\cite{AH2002} confirmed states are called ``kernel states'',
and predicted states are called ``non-kernel states''.
}
Figure \ref{fig:ahfa} on page
\pageref{fig:ahfa} shows the AHFA for
the grammar
\[ \begin{array}{lcl}
  \V{S}^\prime&\de&\V{S}  \\
  \V{S}&\de&\V{A}\cat\V{B} \\
  \V{A}&\de&\V{B} \\
  \V{A}&\de&\V{x}\cat\V{a} \\
  \V{B}&\de&\V{x}\cat\V{b} \qquad .
\end{array} \]

\section{Nihilist normal form}
\label{sec:nihilist-normal-form}

But, with the use of AHFA's, \cite{AH2002} was not
quite done.
Different sequences of null and symbol transitions
could lead to the same state,
making it
difficult to determine which symbol instances
were nulled.
Knowing which of the symbol instances are nulled
is often essential for the semantics.

To solve this, \cite{AH2002} introduced
Nihilist Normal Form (NNF).
NNF is a grammar rewrite that ``factors'' rules with
proper nullables into multiple rules.
As an example,
we consider the grammar
\[ \begin{array}{lcl}
  \V{S}^\prime&\de&\V{S}  \\
  \V{S}&\de&\V{A}\cat\V{A} \\
  \V{A}&\de&\V{a} \\
  \V{A}&\de& \qquad .
\end{array} \]
NNF factors this grammar
by introducing a new nulling symbol, \V{Ae},
and rewriting it as
\[ \begin{array}{lcl}
  \V{S}^\prime&\de&\V{S}  \\
  \V{S}&\de&\V{Ae}\cat\V{Ae} \\
  \V{S}&\de&\V{Ae}\cat\V{A} \\
  \V{S}&\de&\V{A}\cat\V{Ae} \\
  \V{S}&\de&\V{A}\cat\V{A} \\
  \V{A}&\de&\V{a} \\
  \V{Ae}&\de& \qquad .
\end{array} \]
We see that a rule that contains 2 proper nullables
is factored into 4 new rules.
Each rule is factored into \order{2\expon\V{pn}}
new rules,
where \V{pn} is the number of proper nullables in the rule.
This overhead is a grammar constant, and \cite{AH2002} found it tolerable.
We will return to this matter
in Section \ref{sec:the-chaf-rewrite}
on page \pageref{sec:the-chaf-rewrite}.

\section{The pitfall of counting Earley items}
\label{sec:pitfall-of-counting-earley-items}

It is now standard to analyze algorithms, by assigning resource (time or space)
to operations, and looking at the results in asymptotic terms.
The focus on asymptotic terms has the disadvantage of ignoring ``hidden constants'',
which can have a real effect in practice,
but this is far outweighed by the advantage:
Results in asymptotic terms hold up well when
the environment or hardware changes,
which in the computer field they do rapidly.

In analyzing parsing,
results are sometimes reported in terms of the size of
the grammar.
But most often,
the grammar is treated as constant,
and complexity is reported as a function of the input length.

When analyzing the complexity of parsers,
evaluation is usually ignored,
even though in practice an application almost
always wants, not just to parse,
but to evaluate the result of the parse.
From the point of view of the typical application,
evaluation is a necessary part of it,
and the only relevant cost is the combined
cost of both parsing and evaluation.

Applications also usually want run-time features.
But the impact of algorithms on
tracing, debugging, and run-time events
is almost always ignored completely.

\begin{MYsloppy}
Analysis of Earley parsing,
going back to the Earley's original work~\cite{Earley1970},
uses an ``accounting technique'' to calculate complexity.
Operations are not counted directly.
Instead a complex accounting assigns the resource
for every operation to
an attempt to add an Earley item,
taking care that each attempt is assigned
no more than
a constant amount of resource.
The resource consumed by the Earley algorithm is then
determined by counting the number of attempts to create Earley items.
For an unambiguous grammar,
the number of attempts to create Earley items
will be the same as the number of Earley items.
\end{MYsloppy}

Aycock and Horspool~\cite{AH2002}
reported no complexity improvements in
asymptotic terms.
Instead \cite{AH2002} claimed that,
because of its use of AHFA states,
its algorithm produced fewer Earley items,
reducing constants,
and that this reduction would be significant,
even though it was hidden by asymptotic analysis.

In assessing whether \cite{AH2002} was indeed faster,
we need to look for any additional overheads
the use of AHFA states might impose.
These additional overheads would be among costs hidden
by asymptotic analysis,
but when the claimed improvement is only in the
``hidden constants'',
they come into play,
and can reduce or eliminate any gains.

\section{AHFA states are not disjoint}
\label{sec:ahfa-states-are-not-disjoint}

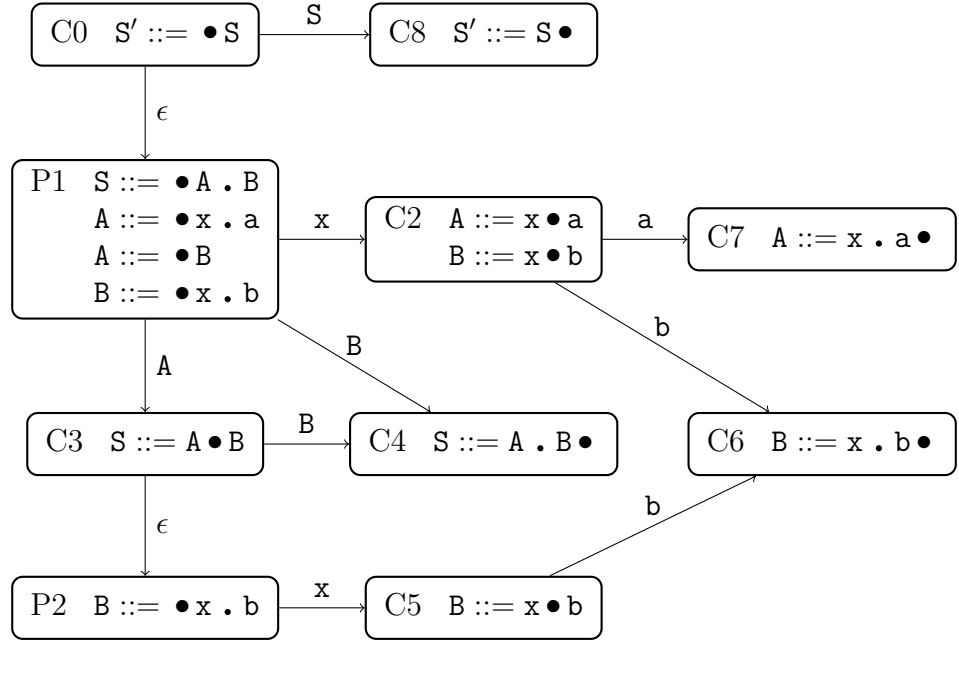
\begin{figure}
\hrule
\caption{\label{fig:ahfa}
    AHFA example
}
\vspace{.5ex}
\hrule
\vspace{.5\baselineskip}
\begin{tikzpicture}

 \node[state,
     anchor=center,
 ] (C0)
 {\begin{tabular}{ll}
   C0&$\V{S}^\prime \de \mydot \V{S}$
   \end{tabular}};

 \node[state,
     anchor=center,
     right of=C0,
     node distance=4.5cm] (C8)
 {\begin{tabular}{ll}
   C8&$\V{S}^\prime \de \V{S} \mydot$
   \end{tabular}};

 \node[state,
     anchor=center,
     below of=C0,
     node distance=15ex] (P1)
 {\begin{tabular}{ll}
   P1&$\V{S} \de \mydot \V{A}\cat\V{B}$ \\
   &$\V{A} \de \mydot \V{x}\cat\V{a}$ \\
   &$\V{A} \de \mydot \V{B}$ \\
   &$\V{B} \de \mydot \V{x}\cat\V{b}$
   \end{tabular}};

 \node[state,
     anchor=center,
     right of=P1,
     node distance=4.5cm] (C2)
 {\begin{tabular}{ll}
   C2&$\V{A} \de \V{x}\mydot\V{a}$ \\
   &$\V{B} \de \V{x}\mydot\V{b}$
   \end{tabular}};

 \node[state,
     anchor=center,
     below of=P1,
     node distance=15ex] (C3)
 {\begin{tabular}{ll}
   C3&$\V{S} \de \V{A}\mydot\V{B}$
   \end{tabular}};

 \node[state,
     anchor=center,
     right of=C3,
     node distance=4.5cm] (C4)
 {\begin{tabular}{ll}
   C4&$\V{S} \de \V{A}\cat\V{B}\mydot$
   \end{tabular}};

 \node[state,
     anchor=center,
     below of=C3,
     node distance=12ex] (P2)
 {\begin{tabular}{ll}
   P2&$\V{B} \de \mydot \V{x}\cat\V{b}$
   \end{tabular}};

 \node[state,
     anchor=center,
     right of=P2,
     node distance=4.5cm] (C5)
 {\begin{tabular}{ll}
   C5&$\V{B} \de \V{x}\mydot\V{b}$
   \end{tabular}};

 \node[state,
     anchor=center,
     right of=C4,
     node distance=4.5cm] (C6)
 {\begin{tabular}{ll}
   C6&$\V{B} \de \V{x}\cat\V{b}\mydot$
   \end{tabular}};

 \node[state,
     anchor=center,
     right of=C2,
     node distance=4.5cm] (C7)
 {\begin{tabular}{ll}
   C7&$\V{A} \de \V{x}\cat\V{a}\mydot$
   \end{tabular}};

  \path[->] (C0) edge node[anchor=left,right] {$\epsilon$} (P1) ;
  \path[->] (C0) edge node[anchor=south,above] {$\V{S}$} (C8) ;
  \path[->] (P1) edge node[anchor=south,above] {$\V{x}$} (C2) ;
  \path[->] (P1) edge node[anchor=left,right] {$\V{A}$} (C3) ;
  \path[->] (P1) edge node[anchor=south,above] {$\V{B}$} (C4) ;
  \path[->] (C3) edge node[anchor=left,right] {$\epsilon$} (P2) ;
  \path[->] (C3) edge node[anchor=south,above] {$\V{B}$} (C4) ;
  \path[->] (P2) edge node[anchor=south,above] {$\V{x}$} (C5) ;
  \path[->] (C2) edge node[anchor=south,above] {$\V{a}$} (C7) ;
  \path[->] (C2) edge node[anchor=south,above] {$\V{b}$} (C6) ;
  \path[->] (C5) edge node[anchor=south,above] {$\V{b}$} (C6) ;
\end{tikzpicture}
\vspace{.5\baselineskip}
\hrule
\end{figure}

The states of an AHFA
are not a partition of the dotted
rules --
a single dotted rule can occur
in more than one AHFA state.
For example, in the AHFA of Figure
\ref{fig:ahfa}, the
dotted rule
\[ \V{B} \de \V{x}\cat\V{b}\mydot \]
occurs in AHFA states \V{C2} and \V{C5}.

Aycock and Horspool do
not explicitly mention this issue.
On \cite[p. 626]{AH2002}, they state that
\begin{quote}
we have
not changed the underlying time complexity of Earley’s
algorithm. In the worst case, each split $\epsilon$-DFA state would
contain a single item, effectively reducing it to Earley’s
original algorithm. Having said this, we are not aware of
any practical example where this occurs.
\end{quote}

\cite{AH2002} does not present a proof that in the worst case
the number of Earley items in their algorithm is the same as
Earley's.
The count of dotted rules does not directly set an upper limit on
the AHFA states,
since the same dotted rule can occur in more than one AHFA.
We were unable to find a proof for their worst case,
but we also could not find a counter-example.

\section{Problems with the AHFA}
\label{sec:problems-with-the-ahfa}

\subsection{Difficulty in debugging and tracing}
\label{sec:difficulty-in-debugging-and-tracing}

During the years that Marpa used \type{AHEM}s,
we encountered a number of disadvantages.
A first was in debugging grammars.

Earley items
are an intuitive way to view the state of a parse.
Each Earley item states a rule,
the amount of progress within the rule
(indicated by the dot),
the location of the dot within the input string,
and the location within the input string
where recognition of the rule began.

Use of AHFA states added a layer of abstraction over this.
We were faced with two choices ---
require a user of the Marpa algorithm to deal with the AHFA states
directly,
or translate them for the user into dotted rules.
Translating the AHFA states into dotted rules
took the user away from the actual workings
of the algorithm.
And the translation was not a simple one.
Since two AHFA states might contain the same dotted rule,
the relationship between AHFA states and dotted rules is
many-to-many.

\subsection{Increased code complexity}
\label{sec:increased-code-complexity}

Most Marpa users,
if they were simply trying to get a grammar working,
could ignore the internal representation,
focus on the dotted rules,
and act as if the AHFA states did not exist.
But a user might, for example,
be concerned with efficiency.
In that case a trace that translated multiple \type{AHEM}s
into a different count of \type{EIM}s might give a misleading impression.

For the programmer working in the Marpa code itself,
dealing with the AHFA states was unavoidable.
Evaluation required doing the many-to-many translation
from AHFA states and dotted rules.
Undoing the potential duplication of dotted rules in AHFA states
is an overhead that cut into the speedups promised by AHFA states,
and the overhead of a duplication check had to be incurred for every \type{AHEM},
whether it contained a duplicated dotted rule or not.

Translation often needed to be done at run-time.
We have mentioned the case of tracing or debugging.
Marpa also has a run-time event mechanism and,
to be practical,
the events in this mechanism needed to be presented to the user in terms
of dotted rules, not AHFA states.
Other run-time features have been contemplated for Marpa,
and the complexity of translating to and from AHFA states
posed a real obstacle to their implementation.

\begin{MYsloppy}
\subsection{Increased theoretical complexity}
\label{sec:increased-theoretical-complexity}

The use of \type{AHEM} items made
the initial version
of Marpa more difficult to describe,
and to analyze theoretically.
Additional theoretical complexity is not just
a theoretical issue.
In developing Marpa, proofs are often needed.
Marpa is written in CWEB~\cite{CWEB2022}
for this reason.
In Marpa, theoretical complexity is an obstacle
to new features, and to maintainance.
\end{MYsloppy}

Earley~\cite{Earley1970}, in his proofs, could rely on each Earley item
representing one dotted rule.
For analyzing Marpa with \type{AHEM}s,
since we did not have a proof that the count of Earley items
in~\cite{Earley1970}
was the worst case for~\cite{AH2002},
we felt unable to do so.
But we did note that the number of AHFA
states is a constant which depends on the grammar,
and that in the worst case a dotted rule is in every AHFA state,
so that the number of occurrences of dotted rules in AHFA states cannot
exceed the number of AHFA states.
(In practice, of course, the number of dotted
rule occurrences will be far less.)
With this assumption,
and the appropriate changes in,
and additions to,
the proofs in
\cite{Earley1970}~and~\cite{Leo1991},
we were able to claim the same asymptotic results
as \cite{Earley1970}~and~\cite{Leo1991}.

\subsection{No noticeable performance improvement}

Given the disadvantages mentioned in Sections
\ref{sec:difficulty-in-debugging-and-tracing}
(page \pageref{sec:difficulty-in-debugging-and-tracing}),
\ref{sec:increased-code-complexity}
(page \pageref{sec:increased-code-complexity}), and
\ref{sec:increased-theoretical-complexity}
(page \pageref{sec:increased-theoretical-complexity}),
in 2014 we decided to explore the possibility of
backing out the use of AHFA states in Marpa.
We would have been willing
to pay a price in performance.

To explore the price,
we counted the size of AHFA states for practical grammars.
Our results are reported in an appendix
starting on~page~\pageref{sec:statistics-on-ahfa-states}.
The relevant finding for our purposes was that,
in practice,
confirmed AHFA states tended to have very few dotted rules,
usually only one.
The average number of dotted rules
in confirmed AHFA states
was considerably less than 2.

This made some sense, since the motivation in~\cite{AH2002}
for AHFA's was memoizing the transitive closure of predictions.
We noted that predictions are very easy to memoize.%
\label{text:prediction-optimization}
For every prediction, call it \Veim{p}, in the Earley set at location \Vloc{i},
we have $\Pos{\V{p}} = 0$
and $\Origin{\V{p}} = \Vloc{i}$,
so that $\Rule{\Veim{p}}$
uniquely identifies \Veim{p}.

As an example of a possible optimization,
since predictions in each Earley set can be uniquely identified by their rule,
if the rules are numbered,
the predictions in each Earley set can be represented as a bit map.
The transitive closure of the predictions for each symbol
could be precomputed as a bit mask, and the bit masks or'ed into the bit map.
The bit map could be retained as the representation of the predictions,
or converted into a list of integers.

In converting away from AHFA states,
we did not use a bit map for adding predictions.
Instead we decided to start with
a simple implementation of prediction,
one which will be described in
Section~\ref{sec:creating-predictions}
on page~\pageref{sec:creating-predictions}.
We converted Marpa incrementally, using test-driven development.
After each change we timed our test suite,
watching for changes in running speed.
The tests in our test suite usually tested not just parsing,
but evaluation.
Some of the tests were also of
run-time event-processing,
and tracing.

Running speed for the test suite remained,
within the precision our measurement allowed,
constant during the changeover.
Our tests of speed were not benchmarks of the typical kind,
which control the environment and narrow the focus,
and which can produce measurements that are in some sense more precise
than the ones we used.
Our tests, instead, reflected the range of circumstances under which
Marpa is likely to be used.
This produced a less precise metric, but one which was more important to us.

The speed measurements from our test suite,
combined with our asymptotic results,
the theoretical indicators that AHFA states were not necessarily
a clear win,
and our statistics on AHFA states,
gave us confidence that
we had either not paid any performance price
for the changeover,
or that the price paid was a very acceptable one.

\section{Marpa's approach to nullable symbols}
\label{sec:marpa-s-approach-to-nullable-symbols}

\subsection{The pitfall of rewrites}

As we saw in Section~\ref{sec:nihilist-normal-form}
on page \pageref{sec:nihilist-normal-form},
Aycock and Horspool, as part of the algorithm described
in~\cite{AH2002},
used a grammar rewrite.
There are many grammar rewrites in the parsing literature
to ``solve'' problems.
For example, if a parser has trouble with left recursion,
a rewrite can convert its grammars to use right recursion.
This rewrite solves the problem in the sense that the parser
running the rewritten grammar
recognizes the same language as the parser does
when running the pre-rewrite grammar.

However, ``language'' has a very non-intuitive meaning in
parsing theory.\footnote{%
    This eccentric use of the term ``language'' in parsing theory,
    perhaps surprisingly,
    is a holdover from a view that was once
    dominant in linguistics.
    The entries in
    my timeline of parsing history~\cite{Timeline}
    for the years from 1929 to 1956
    describe this strange subplot of intellectual history.
}
In parsing theory, a language is a set of strings,
and is not associated with a semantics.

For this reason,
rewrites have been very lightly used for practical parsing.
Practical parsing almost always requires an evaluation phase.
Even if a complex rewrite recognizes the same language,
in the parsing theory sense of the word ``language'',
the application usually also needs to duplicate the semantics associated
with every sentence of the grammar's language.
Many rewrites in the parsing literature
preserve the ``language'',
but erase the relationship between the rules
of the grammar and their semantics,
or at least make that relationship so complex that the benefit
of the rewrite is lost.
For a rewrite to be useful for Marpa's purposes,
the rewritten grammar must come with
semantic processing that duplicates the semantics
of the pre-rewrite grammar.

\subsection{Describing semantics}

For the purposes of this paper,
we will describe semantics using
a simplified version of the
semantics implemented in Marpa.
Marpa's tables are converted into
a parse tree,
which has rule nodes,
and nodes for terminals.
These nodes are evaluated recursively
and the value of the top node
is considered to be the value of the parse.

The semantics of a rule node can be treated as a impure function,
\[ \VFA{semRule}{$\V{parse}, \V{child1}, \V{child2}, \dotsc, \V{childN}$}, \]
where \( \V{child1}, \V{child2}, \dotsc, \V{childN} \)
are the values of the rule instance's child nodes.
\label{text:parse-argument}%
The parse argument, \V{parse}, gives \Fname{semRule}
access to all the data of the parse,
including the grammar,
and the parse location of the rule instance.

Our semantic functions are impure functions ---
they have side effects.
For instance they may build symbol tables.
For our purposes in this paper,
this will not be a problem.

Nulled symbol instances are always associated with
empty rules and their semantics comes from their rule.
Nulled symbol instances may be regarded
as rules with no children.

Every terminal symbol has
an impure semantic function associated with it:
\[ \VFA{semTerm}{\V{parse}, \V{value}}. \]
Here \V{parse} is as for the rule semantics,
and \V{value} is the ``token value'', a value that the application associates
with each token as it is read.

A rule has pass-through semantics if it simply ``passes through''
the value of its child node.
More precisely, a rule node has pass-through semantics iff
it has exactly one child,
and the semantic function is a pure function,
call it \V{semPass},
where
\[ \VFA{semPass}{\V{parse}, \V{child}} = \V{child}. \]

One very common rewrite is ``augmenting the grammar''
with an accept rule.
Many parsers augment their grammars with an accept rule.
Marpa is one of them
(see equation \ref{eq:accept-rule} on page \pageref{eq:accept-rule}).
As an example, a new accept rule might be
\[ \Vrule{accept} = \tuple{ \Vsym{accept} \de \Vsym{start} }, \]
where \Vsym{accept} is a new symbol,
and \V{start} is the start symbol of the original grammar.
We can ensure that the semantics of the augmented grammar duplicate
those of the original grammar by giving the
accept rule pass-through semantics.

In the case of Aycock and Horspool's NNF rules,
ensuring that the rewritten grammar
duplicated the semantics of the original was slightly more complex,
but it was clear that it could be easily and efficiently done.

\subsection{The CHAF rewrite}
\label{sec:the-chaf-rewrite}

Marpa never implemented the NNF rules in the form as
described in~\cite{AH2002}.
As we have mentioned,
the NNF rewrite creates \order{2\expon\V{pn}} new rules
for every rule \V{r} in the original grammar,
where \V{pn} is the count of proper nullables in \V{r}.
This is an overhead which is a constant depending on
the grammar, but
we regarded it as, potentially,
a serious impediment.
For example, the MYSQL 8.0 select statement has
19 optional arguments\footnote{%
    See ``13.2.13 SELECT Statement'' in the MYSQL 8.0 Reference
    Manual~\cite{MYSQL8.0}.
    The direct link is
    \url{https://dev.mysql.com/doc/refman/8.0/en/select.html}.
},
so that the rule for the select statement,
when rewritten into NNF,
would become 524,288 rules.

Once the exponential explosion of NNF rules is considered a problem,
a simple fix suggests itself.
Every rule can be split up,
after the fashion of Chomsky Normal Form (CNF),
into rules which contain no more than 2 proper nullables.
The newly created LHS symbols may also be properly nullable,
but even in the worst case,
the CNF-style split-up results in fewer than \V{pn} rules.
Since each of the smaller rules contains at most 2 proper nullables,
NNF factors it into
no more than 4 rules.
The total number rules of created will therefore be linear:
\[ \order{\V{pn}\mult 4} = \order{\V{pn}}. \]

We called this hybrid of NNF and CNF,
Chomsky-Horspool-Aycock Form (CHAF).
In principle the CHAF rewrite is not difficult,
and it could be left to the user.
But there are a lot of corner cases,
and Marpa automatically applies the CHAF rewrite to all rules
with proper nullables.

For an example of the CHAF rewrite, we will apply a CHAF
rewrite to the augmented grammar\footnote{%
    This is adopted from Aycock and Horspool's main example grammar, as given
    in their Figure 2~\cite[p. 621]{AH2002}.
}
\[ \begin{array}{lcl}
  \V{S}^\prime&\de&\V{S}  \\
  \V{S}&\de&\V{A}\cat\V{A}\cat\V{A}\cat\V{A} \\
  \V{A}&\de&\V{a} \\
  \V{A}&\de&
\end{array} \]
to produce the grammar in
Figure~\ref{fig:chaf}
on page~\pageref{fig:chaf}.
In Figure~\ref{fig:chaf}, the left hand column shows
the rules of the rewritten grammar,
and the right hand column shows the semantics for the new rules.

\begin{figure}
\hrule
\caption{\label{fig:chaf}
    CHAF example
}
\vspace{.5ex}
\hrule
\vspace{.5\baselineskip}
\begin{tabular}{ll}
\multicolumn{1}{c}{Rule} &
\multicolumn{1}{c}{Semantics} \\
\vspace{1.5\baselineskip}
  $ \V{S}^\prime \de \V{S} $ & Pass through \\
  $ \V{S} \de \V{A}\cat\V{S1} $ & CHAF head \\
  $ \V{S} \de \V{A}\cat\V{Ae}\cat\V{Ae}\cat\V{Ae} \qquad $ &
     From pre-rewrite $\V{S} \de \V{A}\cat\V{A}\cat\V{A}\cat\V{A} $ \\
  $ \V{S} \de \V{Ae}\cat\V{S1} $ & CHAF head \\
  $ \V{S1} \de \V{A}\cat\V{S2} $ & CHAF inner \\
  $ \V{S1} \de \V{A}\cat\V{Ae}\cat\V{Ae} $ & CHAF tail \\
  $ \V{S1} \de \V{Ae}\cat\V{S2} $ & CHAF inner \\
  $ \V{S2} \de \V{A}\cat\V{A} $ & CHAF tail \\
  $ \V{S2} \de \V{A}\cat\V{Ae} $ & CHAF tail \\
  $ \V{S2} \de \V{Ae}\cat\V{A} $ & CHAF tail \\
  $ \V{A} \de \V{a} $ & From pre-rewrite $ \V{A} \de \V{a} $ \\
  $ \V{Ae} \de $ & From pre-rewrite $ \V{A} \de $
\end{tabular}
\vspace{.5\baselineskip}
\hrule
\end{figure}

For several of the rules
in Figure~\ref{fig:chaf},
the rewrite is trivial,
and the semantics are exactly those of a pre-rewrite rule.
The accept rule retains its pass-through semantics.

The rules with CHAF semantics
in Figure~\ref{fig:chaf} also use an array, call it \V{childV},
to accumulate the child values of the pre-rewrite rule.
Since the RHS of the pre-rewrite rule had 4 symbols, $\Vsize{childV} = 4$.
In the counting of arguments below, we recall that the first argument
of every semantic function is the \V{parse} argument
(described on page \pageref{text:parse-argument})
so that the child values are passed as the second through last arguments.

\Needspace{10\baselineskip}
If the rule has ``CHAF tail'' semantics,
the semantics function does the following, in sequence:
\begin{itemize}
\item Creates the \V{childV} array.
\item Populates the \V{childV} array,
  from right to left,
  with the semantics function's arguments from last to second.
  For example, the semantic function for the $ \V{S2} \de \V{Ae}\cat\V{A} $ rule,
  call it \Fname{semS2c},
  will write \Fname{semS2c}'s last argument (for the symbol \V{A}) into $\Velement{childV}{3}$,
  and \Fname{semS2c}'s second argument (for the symbol \V{Ae}) into $\Velement{childV}{2}$.
\item Returns \V{childV} as its value.
\end{itemize}

If the rule has ``CHAF inner'' semantics,
its semantics function does the following, in sequence:
\begin{itemize}
\item Locates the \V{childV} array, which will be the last argument
of the semantics function.
\item
  Populates the unpopulated elements of the \V{childV} array
  from right to left,
  with the semantics function's arguments from next-to-last to second.
  For example, the semantics function for the $ \V{S1} \de \V{Ae}\cat\V{S2} $ rule,
  call it \Fname{semS1b},
  will write \Fname{semS1b}'s next-to-last argument
  into $\Velement{childV}{1}$.
  The next-to-last argument
  of \Fname{semS1b}
  is also \Fname{semS1b}'s second argument
  and comes from the symbol \V{Ae}.
\item Returns \V{childV}.
\end{itemize}

For purpose of describing the ``CHAF head'' semantics,
the rule
\begin{equation}
\label{eq:pre-rewrite-rule}
  \V{S} \de \V{A}\cat\V{A}\cat\V{A}\cat\V{A}
\end{equation}
of the pre-rewrite grammar is the ``pre-rewrite rule''.
If the rule has ``CHAF head'' semantics,
its semantics function does the following, in sequence:
\begin{itemize}
\item Locates the \V{childV} array, which will be the last argument
of the semantics function.
\item
  Populates the unpopulated elements of the \V{childV} array
  from right to left,
  with the semantics function's arguments from next-to-last to second.
  For example, the semantics function for the
  $ \V{S} \de \V{Ae}\cat\V{S1} $ rule,
  call it \Fname{semSa},
  will write \Fname{semSa}'s next-to-last argument
  into $\Velement{childV}{0}$.
  The next-to-last argument
  of \Fname{semSa}
  is also \Fname{semSa}'s second argument
  and comes from the symbol \V{Ae}.
\item Sets a temporary variable, call it \V{v},
    to the semantics function of
    the pre-rewrite rule~\eqref{eq:pre-rewrite-rule}
    applied to the elements of \V{childV},
    so that
    \[ \begin{gathered}
       \V{v} = \Fname{semOrig}(\V{parse}, \Velement{childV}{0}, \Velement{childV}{1}, \\
	 \qquad \qquad \Velement{childV}{2}, \Velement{childV}{3}),
    \end{gathered} \]
    where \Fname{semOrig} is the semantics function for \eqref{eq:pre-rewrite-rule}.
\item Releases the memory for \V{childV}, depending on the memory management model.
\item Returns \V{v}.
\end{itemize}

The ``CHAF head'', ``CHAF inner'', and ``CHAF tail'' semantic functions are
all pure functions.
In all of the CHAF semantic routines,
which argument goes into which slot of \V{childV}
is known once the grammar rewrite is complete.
It does not have to be determined at run-time.

\begin{MYsloppy}
The semantic functions just given for CHAF rules are intended
to demonstrate that the semantics of the CHAF rewrite can be
implemented efficiently.
They do not reflect the implementation of Marpa~\cite{Marpa-R2}.
In~\cite{Marpa-R2},
all the CHAF rules share a unified logic,
which is driven by parameters that are set at CHAF rewrite time for each CHAF rule.
\cite{Marpa-R2}~does not use a \V{childV} array.
Instead it keeps the child values on a stack.
\end{MYsloppy}

\subsection{Eliminating nulling symbols}

With the CHAF rewrite,
we have eliminated proper nullables,
so the symbols and rules in the grammar are either nulling
or non-nullable.
We now make the further observation that
nulling symbols and nulling rules do not
show any trace in the input.
They can be eliminated.

We will call a grammar before the elimination of nulling symbols
the ``nulling-present'' grammar.
We will call the grammar after elimination of nulling symbols
the ``nulling-free'' grammar.
The nulling-present grammar and the nulling-free grammar accept the same language.

To duplicate the semantics of the nulling-present grammar,
we keep a database of ``nulling markup'',
which tracks the locations
in the nulling-free rules
where nulling symbols were eliminated.
We can go back and forth quickly between a nulling-free rule
and its nulling markup, on one hand,
and a nulling-present rule,
on the other hand.

The semantics of the nulling-present grammar can easily be replicated
at evaluation time using the nulling markup.
This can also be done efficiently at run-time,
so that Marpa's recognizer runs using a nulling-free
grammar,
but its tracing and event-generation is done in terms
of the nulling-present grammar.

For the elimination of nulling symbols and rules
to succeed, several corner cases must be dealt with.
These are trivial grammars, trivial parses,
nulling subforests,
and duplicated rules.
Trivial grammars are grammars which recognize only the
empty string.
Trivial parses are parses of the empty string.
The Marpa implementation deals with
trivial grammars and trivial parses
by treating them as special cases.

The elimination of nulling rules can prune entire subforests from a parse.
Since these subforests are constants which depend on the grammar,
we could,
in the same way that we restore the nulling symbols during evaluation and tracing,
restore these pruned subforests during evaluation and tracing.

In practice, we found that users are not interested in the
``semantics of nothing''.
For evaluation purposes,
Marpa's current implementation prunes nulled subforests back to
their topmost symbol.
Applications are free to restore the semantics of the pruned
subforests, but none have chosen to do so,
to our knowledge.

Users do want run-time events to take pruned symbols into account.
The Marpa implementation's run-time event feature
generates events in terms of the nulling-present grammar.
The effect of nulling symbols on each event is a constant which depends on
the grammar,
and run-time nulling-awareness is very efficient.

Elimination of nulling symbols also raises an issue
of duplicated rules.
Two rules are traditionally considered identical if their
LHS and RHS are identical.
But elimination of nulling symbols can turn distinct rules
into rules which are identical by this definition.
For example,
in the grammar of our Figure~\ref{fig:chaf}
on page \pageref{fig:chaf},
the two rules
\begin{equation*}
  \tuple{\V{S2} \de \V{A}\cat\V{Ae}}
\end{equation*}
and
\begin{equation*}
  \tuple{\V{S2} \de \V{Ae}\cat\V{A}},
\end{equation*}
when the nulling symbol \V{Ae} is eliminated,
both become the rule
\begin{equation}
\label{eq:dup-rule}
\tuple{\V{S2} \de \V{A}}.
\end{equation}

In Marpa's internals,
nulling markup is taken into account
in identifying a rule.
In effect, the Marpa implementation has two distinct rules of
the form in \eqref{eq:dup-rule},
which are distinguished by their nulling markup.

The duplication of rules which differ only in their nulling markup could
also have been avoided with a rewrite.
If, in the grammar of our Figure~\ref{fig:chaf},
the rules with \V{S2} on the LHS are removed and replaced
with
\[ \begin{array}{lcl}
  \V{S2}&\de&\V{S2a} \\
  \V{S2}&\de&\V{S2b} \\
  \V{S2a}&\de&\V{A}\cat\V{A} \\
  \V{S2a}&\de&\V{A}\cat\V{Ae} \\
  \V{S2b}&\de&\V{Ae}\cat\V{A} \qquad ,
\end{array} \]
then the rules are differentiated by their LHS's,
and duplication is avoided.

\subsection{Creating predictions}
\label{sec:creating-predictions}

We now are in a position to return to the issue
of creating predictions.
We recall from page \pageref{text:prediction-completion-cycle}
in Section \ref{text:prediction-completion-cycle} that,
in traditional Earley implementations,
predictions could give rise to completions,
which in turn could cause new predictions.

With the elimination of nullable symbols,
a prediction can no longer be a completion in the same Earley set,
and cannot give rise to a completion.
This allows Marpa to
complete each Earley set, call it \V{es},
in distinct phases:
\begin{itemize}
\item Marpa first creates \V{es} and adds all scanned items to it.
\item Marpa next adds all reductions to \V{es}.
\item Finally, Marpa adds predictions to \V{es}.
\end{itemize}
Unlike many Earley implementations,
including the one in~\cite{AH2002},
Marpa works on only one Earley set at a time.
This means there is a point in the running of the Marpa
parse engine, where work on the Earley set at location \Vloc{i}
is complete,
but work on the Earley set at \Vloc{i}+1 has yet to be started.
At this point the parse is ``left-eidetic'' ---
fully aware of everything that has taken place up to the current
location,
but not yet committed to any actions after the current location.

Left-eideticism is very useful for debugging and tracing.
Error detection in Marpa is efficient and accurate enough that
``soft errors'' can be deliberately exploited as a parsing technique.
For example, when parsing fails because no acceptable token is supplied,
the application can ask Marpa what tokens would have been acceptable ---
this is known because the full left context of the parse is known.
This technique, described in~\cite{Marpa2023}, is called the ``Ruby Slippers''.

One application of the Ruby Slippers has been to base an extremely liberal
HTML parser~\cite{Marpa-HTML}
on an over-conservative grammar.
The over-conservative HTML grammar
is stricter than the strictest HTML standard
and requires that, among other things, all start tags have matching end tags.
When a parse with the over-conservative grammar fails
the grammar's excessively strict requirements,
\cite{Marpa-HTML}~invents an acceptable token to satisfy it,
allowing the parse to continue.
While the grammar of \cite{Marpa-HTML} is over-conservative,
\cite{Marpa-HTML}
is in operation a completely liberal HTML parser,
one which will parse any input as HTML,
albeit highly defective HTML.

Marpa's method for calculating predictions is simple.
The transitive closure of predictions for each postdot symbol is
precomputed from the grammar.
When creating each Earley set,
once the scanned and reduced Earley items are known,
we make a second pass through them,
looking for their postdot symbols.
For each postdot symbol,
the Earley items in the transitive closure predicted by that postdot symbol
are added to the Earley set in the usual way.

As mentioned earlier
(page
\pageref{text:prediction-optimization}
in Section
\ref{text:prediction-optimization}),
predictions are good candidates for optimization.
Since the Marpa parse engine does not calculate
predictions until all other Earley items in the same Earley set
have been calculated,
the optimization-minded programmer has a full range of options.
Bitmaps or other techniques for the computation of
predicted Earley items
might be faster than the methods Marpa currently uses.
But as we stated, in 2014 we found that our changes did not noticeably
change the speed of Marpa\footnote{%
     Our measurements included evaluation as well as some exercise
     of the run-time facilities.
     In those terms
     it is possible that the 2014 changes sped Marpa up,
     but we do not claim this.}
 and we have left prediction optimization as topic for future research.

\section{Conclusion}
\label{sec:conclusion}

When first released in 2011,
Marpa adopted its approach for dealing with nullable
symbols from Aycock and Horspool~\cite{AH2002}.
The solution of~\cite{AH2002} centered around a
semi-deterministic finite automata,
and converted the Earley algorithm to work in
terms of states of this automata,
rather than dotted rules.

In using and extending Marpa,
we found this solution awkward.
In 2014,
we decided to back the Aycock and Horspool automata out of Marpa.
We did so successfully,
and without a noticeable loss in performance.
But we needed a new approach to nullable symbols.

Fortunately, in the process of studying the
algorithm of~\cite{AH2002},
we discovered that~\cite{AH2002}
contained the key ideas for another solution,
one based on grammar rewrites.
Marpa's current approach to nullable symbols
is an extension of grammar rewriting ideas from~\cite{AH2002}.
For this reason, we continue to consider Marpa to be
a parser in the Aycock-Horspool lineage.

\bibliographystyle{plain}

\Needspace{30\baselineskip}

\appendix

\section{Statistics on AHFA states}
\label{sec:statistics-on-ahfa-states}

\subsection{The test grammars}

For these statistics we used
a standards-quality C grammar,
a prototype Perl grammar,
and a set of HTML grammars that were generated by
a fully useable HTML parser.
The C grammar was
for ISO ANSI C 2011, as used by~\cite{MarpaX-C-AST}.
The prototype Perl grammar was a grammar for a large subset of
Perl.\footnote{%
\raggedright
The Perl grammar is part of \cite{Marpa-R2}, and
as of March 4, 2023 can be accessed in its Github
repository as \url{https://github.com/jeffreykegler/Marpa--R2/blob/master/cpan/pperl/Marpa/R2/Perl.pm}.
}

The HTML parser\footnote{
  The HTML parser is part of \cite{Marpa-R2}.
  As of March 4, 2023 it can be accessed as
  \url{https://metacpan.org/dist/Marpa-R2/view/html/pod/HTML.pod}.
}
is very liberal,
allows the user to configure it,
and creates a customized HTML grammar
based on its configuration.
The statistics reported are aggregates
over all of the customized HTML grammars.

\subsection{Confirmed states}

\FloatBarrier

\begin{table}[H]
\caption{
  \label{tab:perl-confirmed}
  Confirmed AHFA states for a Perl grammar}
\vspace{1ex}
\begin{tabular}{|r|r|}
\hline
\multicolumn{1}{|c|}
  {\rule{0pt}{2.3ex}
   \rule{0pt}{-.5ex}
   Size}&
  \multicolumn{1}{|c|}{Percent}\\
& \multicolumn{1}{|c|}{of occurrences}\\
\hline
\rule{0pt}{2.3ex}
1&67.05\%\\
2&25.67\%\\
3&2.87\%\\
4&2.68\%\\
5&0.19\%\\
6&0.38\%\\
7&0.19\%\\
8&0.57\%\\
9&0.19\%\\
20&0.19\%\\
\hline
\end{tabular}
\end{table}

\FloatBarrier
\subsubsection{Confirmed states for a Perl grammar}
For confirmed states of the Perl grammar,
we show the counts
in Table \ref{tab:perl-confirmed}
on page \pageref{tab:perl-confirmed}.
They range in size from 1 to 20 items,
but the numbers are heavily skewed toward the low
end.
As can be seen, well over 90\% of the total confirmed states have
just one or two items.
The average size is 1.5235,
and the average of the size squared is 3.9405.
\FloatBarrier

\begin{table}[H]
\caption{
  \label{tab:html-confirmed}
  Confirmed AHFA states for HTML grammars}
\vspace{1ex}
\begin{tabular}{|r|r|}
\hline
\multicolumn{1}{|c|}
  {\rule{0pt}{2.3ex}
   \rule{0pt}{-.5ex}
   Size}&
  \multicolumn{1}{|c|}{Percent}\\
& \multicolumn{1}{|c|}{of occurrences}\\
\hline
\rule{0pt}{2.3ex}
1&80.96\%\\
2&19.04\%\\
\hline
\end{tabular}
\end{table}

\FloatBarrier
\subsubsection{Confirmed states for HTML grammars}
For confirmed states of the HTML grammars,
we show the data in
Table \ref{tab:html-confirmed}
on page \pageref{tab:html-confirmed}.
The average size is 1.1904,
and the average of the size squared is 1.5712.

\FloatBarrier 
\begin{table}[H]
\caption{
  \label{tab:c-confirmed}
  Confirmed AHFA states for a C grammar}
\vspace{1ex}
\begin{tabular}{|r|r|}
\hline
\multicolumn{1}{|c|}
  {\rule{0pt}{2.3ex}
   \rule{0pt}{-.5ex}
   Size}&
  \multicolumn{1}{|c|}{Occurrences}\\
\hline
\rule{0pt}{2.3ex}
1&695\\
2&188\\
3&40\\
4&17\\
5&6\\
6&8\\
7&6\\
8&4\\
9&1\\
10&2\\
12&2\\
15&1\\
\hline
\end{tabular}
\end{table}

\FloatBarrier
For confirmed states of the C grammar,
we show the data
in Table \ref{tab:c-confirmed}
on page \pageref{tab:c-confirmed}.
The confirmed states range in size from 1 to 15 items but again,
the numbers are heavily skewed toward the low
end.
There were 970 confirmed C states.
The average size was 1.52.
The average of the size squared was 3.98.
\FloatBarrier

\subsection{Predicted states.}

The number of predicted states tends to be much more
evenly distributed.
It also tends to be much larger.
For predicted states, because AHFA states of varied
size are common,
we switch from showing the frequency of AHFA states
for each size,
to showing the sizes of the AHFA state by the
frequency of states of that size.

\FloatBarrier
\begin{table}[H]
\caption{
  \label{tab:perl-predicted}
  Predicted AHFA states for a Perl grammar}
\vspace{1ex}
\begin{tabular}{|l|r|}
\hline
\multicolumn{1}{|c|}
  {\rule{0pt}{2.3ex}
   \rule{0pt}{-.5ex}
   Size of AHFA state}&
  \multicolumn{1}{|c|}{Occurrences of states}\\
   &\multicolumn{1}{|c|}{of that size}\\
\hline
\rule{0pt}{2.3ex}%
2&5\\
\hline
\rule{0pt}{2.3ex}%
3, 142&4\\
\hline
\rule{0pt}{2.3ex}%
1, 4&3\\
\hline
\rule{0pt}{2.3ex}%
6, 7, 143&2\\
\hline
\cellbox[3in]{
  5, 64, 71, 77, 79, 81, 83, 85, 88, 90, 98, 100, 102,
  104, 106, 108, 111, 116, 127, 129, 132, 135, 136, 137, 141,
  144, 149, 151, 156, 157, 220, 224, 225
}
  &1\\
\hline
\end{tabular}
\end{table}

\FloatBarrier
\subsubsection{Predicted states for a Perl grammar}

Table \ref{tab:perl-predicted}
on page \pageref{tab:perl-predicted}
shows the data for the predicted states of a Perl grammar.
The number of predicted states in the Perl grammar was 58.
The average size was 83.59 dotted rules.
The average of the size squared was 11356.41.
\FloatBarrier

\begin{table}[H]
\caption{
  \label{tab:html-predicted}
  Predicted AHFA states for HTML grammars}
\vspace{1ex}
\begin{tabular}{|r|r|c|r|r|}
\cline{1-2} \cline{4-5}
\multicolumn{1}{|c|}
  {\rule{0pt}{2.3ex}
   \rule{0pt}{-.5ex}
   Size}&
  \multicolumn{1}{|c|}{Occurrences}&
  \multicolumn{1}{|c|}{}&
\multicolumn{1}{|c|}{Size}&
  \multicolumn{1}{|c|}{Occurrences}\\
\cline{1-2} \cline{4-5}
\rule{0pt}{2.3ex}%
1&95&&
20&190\\
\cline{1-2} \cline{4-5}
\rule{0pt}{2.3ex}%
2&95&&
21&63\\
\cline{1-2} \cline{4-5}
\rule{0pt}{2.3ex}%
4&95&&
22&22\\
\cline{1-2} \cline{4-5}
\rule{0pt}{2.3ex}%
11&181&&
24&8\\
\cline{1-2} \cline{4-5}
\rule{0pt}{2.3ex}%
14&181&&
25&16\\
\cline{1-2} \cline{4-5}
\rule{0pt}{2.3ex}%
15&294&&
26&16\\
\cline{1-2} \cline{4-5}
\rule{0pt}{2.3ex}%
16&112&&
28&2\\
\cline{1-2} \cline{4-5}
\rule{0pt}{2.3ex}%
18&349&&
29&16\\
\cline{1-2} \cline{4-5}
\rule{0pt}{2.3ex}%
19&120\\
\cline{1-2}
\end{tabular}
\end{table}

\FloatBarrier
\subsubsection{Predicted states for HTML grammars}

Data for predicted states of the HTML grammars
is in Table \ref{tab:html-predicted}
on page \pageref{tab:html-predicted}.
The total number of predicted states in the HTML grammars was
1855. Their average size was 14.60. Their average size squared was
250.93.
\FloatBarrier

\begin{table}[H]
\caption{
  \label{tab:c-predicted}
  Predicted AHFA states for a C grammar}
\vspace{1ex}
\begin{tabular}{|l|r|}
\hline
\multicolumn{1}{|c|}
  {\rule{0pt}{2.3ex}
   \rule{0pt}{-.5ex}
   Size of AHFA state}&
  \multicolumn{1}{|c|}{Occurrences of states}\\
   &\multicolumn{1}{|c|}{of that size}\\
\hline
\rule{0pt}{2.3ex}%
2, 3&6\\
\hline
\rule{0pt}{2.3ex}%
8&5\\
\hline
\rule{0pt}{2.3ex}%
4, 90&4\\
\hline
\rule{0pt}{2.3ex}%
6, 11, 31, 47&3\\
\hline
\rule{0pt}{2.3ex}%
\cellbox[3in]{5, 14, 42, 64, 68, 78, 91, 95, 98}
&2\\
\hline
\cellbox[3in]{
1, 7, 9, 12, 15, 17, 18, 19, 21, 22, 25, 28, 29, 33, 34, 36,
37, 40, 43, 44, 45, 46, 52, 53, 54, 57, 58, 61, 65, 66, 69, 72,
74, 76, 80, 81, 86, 87, 89, 94, 96, 97, 99, 102, 105, 108,
115, 117, 119, 123, 125, 127, 144, 149, 150, 154, 181, 219,
222.
}
  &1\\
\hline
\end{tabular}
\end{table}

\FloatBarrier
\subsubsection{Predicted states for a C grammar}

The data for predicted states in a C grammar
is in Table \ref{tab:c-predicted}
on page \pageref{tab:c-predicted}.
The number of predicted states in the C grammar was 114.
The average size was 54.81.
The average size squared was 5361.28.
The sizes of the predicted states for the C grammar were spread from 1
to 222.
\FloatBarrier

\subsection{Completed LHS symbols per AHFA state. }
An AHFA state may contain completions for more than one LHS,
but that is rare in practical use, and the number of completed
LHS symbols in the exceptions remains low.
The very complex Perl AHFA contains 271 states with completions.
Of these 268 have only one completed symbol.
The other three AHFA states complete only two different LHS symbols.
Two states have completions with both
a \texttt{term\_hi} and an \texttt{indirob} on the LHS.
One state has completions for both a
\texttt{sideff} and an \texttt{mexpr}.

Our HTML test grammars make the
same point more strongly.
In my HTML test suite,
every single one
of the 14,782 AHFA states
has only one completed LHS symbol.

\FloatBarrier

\clearpage
\tableofcontents

\end{document}